%% file: main.tex
\documentclass[11pt,a4paper]{article}
\usepackage[hyperref]{acl2019}
\usepackage{times}
\usepackage{latexsym}
\usepackage{graphicx}
\usepackage{booktabs}
\usepackage{tabularx}
\usepackage{makecell}
\usepackage{xcolor}
\usepackage{anyfontsize}

\usepackage{url}

\aclfinalcopy 

\newif\ifcomment
\commentfalse

\ifcomment
\newcommand{\uk}[1]{\textcolor{red}{UK: #1}}
\newcommand{\kc}[1]{\textcolor{blue}{KC: #1}}
\newcommand{\dan}[1]{\textcolor{green}{DJ: #1}}
\newcommand{\lk}[1]{\textcolor{cyan}{LK: #1}}
\else
\newcommand{\uk}[1]{}
\newcommand{\kc}[1]{}
\newcommand{\dan}[1]{}
\newcommand{\lk}[1]{}
\fi

\title{Sample Efficient Text Summarization Using a Single Pre-Trained Transformer}

\author{Urvashi Khandelwal$^\dagger$, Kevin Clark$^\dagger$, Dan Jurafsky$^\dagger$, \L{}ukasz Kaiser$^\ddagger$ \\
  $^\dagger$Computer Science Department, Stanford University\\
  $^\ddagger$Google Brain \\
  {\tt \{urvashik,kevclark,jurafsky\}@stanford.edu} \\
  {\tt lukaszkaiser@google.com}\\}

\date{}

\begin{document}
\maketitle
\begin{abstract}
  \input abstract
\end{abstract}

\input intro
\input methods
\input experiments

\input discussion

\input conclusion

\section*{Acknowledgments}
This work was started and, in part, carried out during the first author's internship at Google Brain.
We thank Ashwin Paranjape and Yuhao Zhang for their thoughtful comments and suggestions.
We gratefully acknowledge support of the DARPA Communicating with Computers (CwC) program under ARO prime contract no. W911NF15-1-0462 and the NSF via grant IIS-1514268.
Kevin is supported by a Google PhD Fellowship.



\bibliography{acl2019}
\bibliographystyle{acl_natbib}

\appendix

\input appendix

\end{document}

%% file: abstract.tex

Language model (LM) pre-training has resulted in impressive performance and sample efficiency on a variety of language understanding tasks.
However, it remains unclear how to best use pre-trained LMs for generation tasks such as abstractive summarization, particularly to enhance sample efficiency.
In these sequence-to-sequence settings, prior work has experimented with loading pre-trained weights into the encoder and/or decoder networks, but used non-pre-trained encoder-decoder attention weights.
We instead  use a pre-trained decoder-only network, where the same Transformer LM both encodes the source and generates the summary.
This ensures that all parameters in the network, including those governing attention over source states, have been pre-trained before the fine-tuning step.
Experiments on the CNN/Daily Mail dataset show that our pre-trained Transformer LM substantially improves over pre-trained Transformer encoder-decoder networks in limited-data settings. 
For instance, it achieves 13.1 ROUGE-2 using only 1\% of the training data ($\sim$3000 examples), while pre-trained encoder-decoder models score 2.3 ROUGE-2.

%% file: intro.tex

\section{Introduction}
\label{sec:intro}

Language model (LM) pre-training has led to impressive results on tasks ranging from text classification to sequence tagging to question answering \citep{dai2015semi,peters2018deep,radford2018improving,devlin2018bert}. 
Particularly striking is the improved sample efficiency achieved by these models \citep{howard2018universal}.
However, it remains unclear how to best utilize pre-trained LMs for generation tasks such as text summarization, and how much sample efficiency gains would still apply given that generation models often require large datasets to perform well.

\begin{figure}[t]
\includegraphics[width=0.48\textwidth]{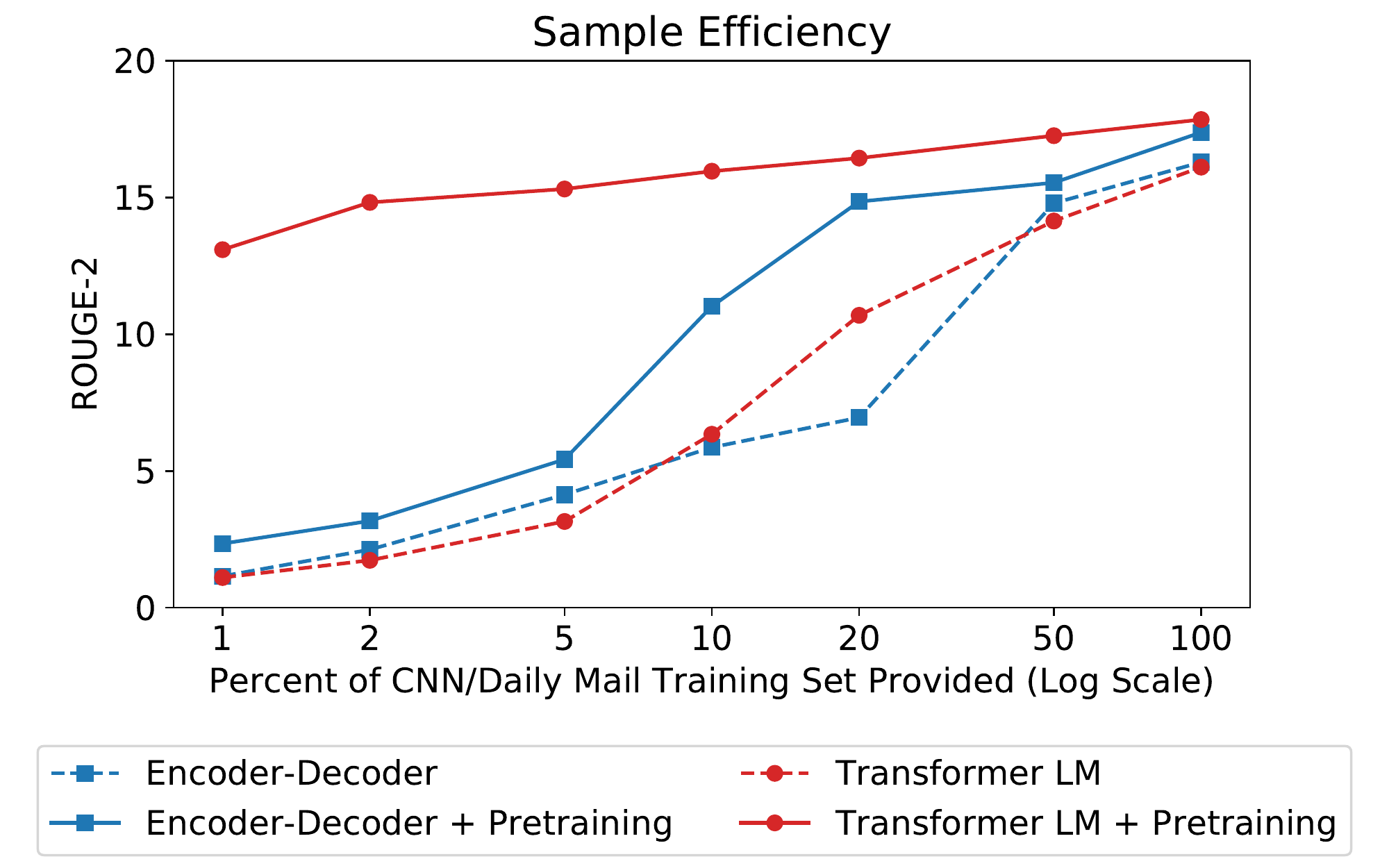}
\caption{A pre-trained Transformer LM is extremely sample efficient, outperforming baselines by over 10 ROUGE-2 points, when fine-tuned on only 1\% data ($\sim$3,000 examples). 
}
\label{fig:sample_efficiency}
\end{figure}

While prior work has explored improving sequence-to-sequence models by incorporating pre-trained weights from LSTM LMs \citep{Glehre2015OnUM,ramachandran2016unsupervised}, their models include many non-pre-trained parameters, such as additional LSTM layers or the weights governing the models' attention mechanisms.
These parameters have to be trained from scratch, which can require 
lots of labeled data.
On the other hand, \citet{radford2019language} have recently trained a large Transformer \citep{vaswani2017attention} language model and applied it to summarization without any fine-tuning (thus adding no non-pre-trained parameters), demonstrating the model's zero-shot abilities.

We explore pre-training a large Transformer language model and fine-tuning it for text summarization, demostrating the model's sample efficiency. 
In order to use pre-trained weights more efficiently, we use a Transformer-based decoder-only network \citep{liu2018generating} during fine-tuning.
This decoder-only network, or Transformer LM, treats summarization as a language modeling task where each example consists of a summary appended to its article.
Rather than using separate encoder and decoder components, a single network is used to both encode the source and generate the target.
Crucially, it includes pre-trained self-attention parameters which are used to attend to both the source and the previously generated target representations.
This approach (1) avoids the redundancy of loading copies of the same pre-trained weights into the encoder and decoder, (2) uses fewer parameters compared to encoder-decoder networks, and most importantly (3) ensures all model weights, including those controlling attention over source states, are pre-trained.


The pre-trained Transformer LM performs competitively on the CNN/Daily Mail dataset,  despite using a simple model without augmentations like a copy mechanism \citep{see2017get} or reinforcement learning \citep{Chen2018FastAS}.
Crucially, this model is extremely sample efficient.
Fine-tuning it on only 1\% data ($\sim$3000 examples), results in a ROUGE-2 score of 13.1, while pre-trained Transformer encoder-decoder networks perform poorly with a ROUGE-2 of 2.3 (see Figure~\ref{fig:sample_efficiency}).
Such a highly sample efficient model opens the door for training summarization models for narrow domains, low-resource languages, or other settings without abundant labeled training data.
In addition, analysis shows that our model is more abstractive than the pointer-generator model \cite{see2017get}.

%% file: methods.tex

\section{Methods}
\label{sec:methods}

In this section, we describe the neural architectures and methods used for training a language model and fine-tuning it for text summarization.

\paragraph{Language Model Pre-training.}
In this study, we train a Transformer \citep{vaswani2017attention} based language model.
Unlike ELMo \citep{peters2018deep}, which trains LMs in both directions, or BERT \citep{devlin2018bert}, which trains a bidirectional word imputation model, we train a unidirectional LM \citep{radford2019language}. This is necessary for initializing an auto-regressive decoder.

\begin{figure}[t]
\includegraphics[width=0.47\textwidth]{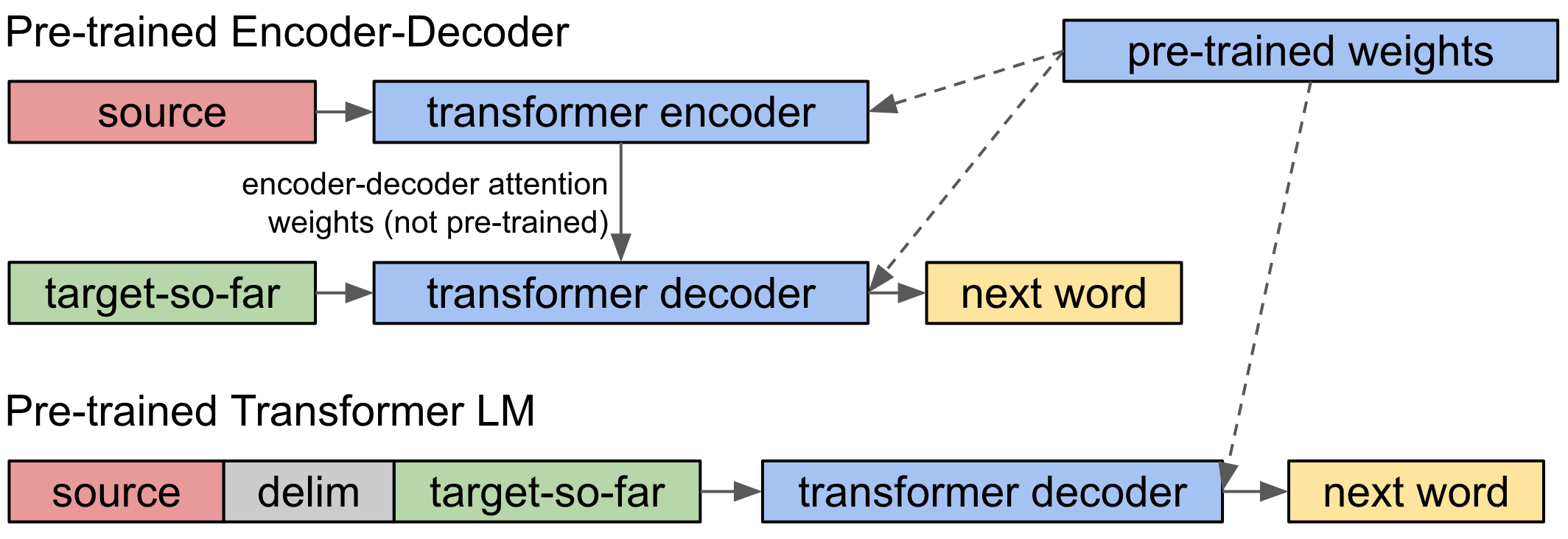}
\caption{Our pre-trained Transformer LM uses the same network to process both the source and target. The encoder-decoder attention parameters, which are not pre-trained, are no longer necessary.}
\label{fig:textsum_as_lm_png}
\vspace{-2mm}
\end{figure}

\paragraph{Encoder-Decoder Baselines.}
We first combine pre-training with standard sequence-to-sequence \cite{sutskever2014sequence} models.   
These consist of a Transformer encoder network that reads the article, a Transformer decoder network that generates the summary, and an encoder-decoder attention mechanism  \cite{bahdanau2014neural} that allows the decoder to attend to encoder states during generation, shown in Figure~\ref{fig:textsum_as_lm_png}.
Both the encoder and decoder use the same  network architecture as the Transformer LM, making it easy to use the pre-trained weights. 
We compare three ways  of  incorporating weights from a pre-trained LM, proposed  by \citet{ramachandran2016unsupervised}: (1) pre-training the encoder only, (2) pre-training the decoder only, and (3) pre-training both. In (3), the encoder-decoder attention parameters are the only ones randomly initialized. 
When fine-tuning these models on summarization data, we follow recent work \citep{howard2018universal,radford2018improving,devlin2018bert} and fine-tune all of the pre-trained as well as non-pre-trained parameters.

\paragraph{Transformer LM.}
We can simplify the encoder-decoder model by casting summarization as a language modeling task \citep{liu2018generating}. 
This is done by appending each target (the summary) to its source (the article), along with a delimiter, and training a Transformer on this reformulated data.
Similar to encoder-decoder models, we only compute loss over the target sequence, as adding loss from the source sequence did not improve performance.
In this setting, a Transformer based decoder-only network is used to process both the source and the target, as shown in Figure~\ref{fig:textsum_as_lm_png}.
The self attention of this decoder-only network, or Transformer LM, is now used to attend to source states as well as the states of already generated targets.
This approach allows for better utilization of pre-trained LMs by removing all non-pre-trained parameters from the model, compared to encoder-decoder models where encoder-decoder attention parameters are not pre-trained.
This is a key benefit of using Transformer LMs over LSTM LMs, which do not include attention mechanisms. 
Using a Transformer LM also avoids the redundancy of loading copies of the pre-trained weights into both the encoder and the decoder.

\paragraph{Model Augmentations.}
While there has been extensive research on model augmentations for text summarization \citep{Glehre2016PointingTU,pasunuru2018multi,li2018improving}, they are often complicated or summarization-specific. 
Therefore, since we want to both clearly isolate the benefits of LM pre-training and experiment with a general model applicable to many domains, we do not augment our simple models. We note, however, that many of these augmentations could be added to our models to further improve performance.

%% file: experiments.tex

\section{Experiments}
\label{sec:exps}

In this section, we describe our data, models, training details, and results. Finally, we discuss the impressive sample efficiency of our model.


\paragraph{Pre-Training Data.} 
While recent prior work \citep{radford2018improving} trains their language model on the Toronto BookCorpus \cite{zhu2015moviebook}, this dataset appears to no longer be publicly available.\footnote{See \url{http://yknzhu.wixsite.com/mbweb}}
Therefore, we instead collected a new 2-billion-word corpus based on Wikipedia called WikiLM.
To facilitate future research on generative pre-training, we make the corpus publicly available.\footnote{Available to download at \url{https://github.com/tensorflow/tensor2tensor}}
\paragraph{Summarization Data.}
We use the non-anonymized CNN/Daily Mail dataset \cite{see2017get}, which involves summarizing news articles into 2-3 sentences. 
Summarization performance is typically evaluated using ROUGE-1 (unigram overlaps), ROUGE-2 (bigram overlaps) and ROUGE-L (subsequence overlaps) \cite{lin2004rouge}.
Although not perfect, these metrics serve as a good first approximation to quantify performance.

\paragraph{Models.} 
For baselines, we evaluate encoder-decoder models with no pre-training as well as the three pre-training strategies discussed in Section ~\ref{sec:methods}.
Using weights from a pre-trained LM constrains the model to be unidirectional and very large, so we report results from a smaller model with a bidirectional encoder to quantify how this affects performance. 
We report results from our single Transformer LM with and without pre-training.
Additionally, we provide scores for a small 4-layer Transformer with a copy mechanism and coverage loss \cite{gehrmann2018bottom}, as a reference for the kind of gains achieved by adding these augmentations.
We also provide zero-shot summarization scores from \citet{radford2019language} which uses a ten times larger pre-trained Transformer language model, but without any fine-tuning, and thus also without introducing non-pre-trained parameters.
Lastly, we provide scores from the state-of-the-art LSTM based model \cite{celikyilmaz2018deep}, which includes a reinforcement learning objective.

\begin{table}
    \centering
    \small
    \setlength{\tabcolsep}{0.4em}
    \begin{tabular}{lccc}
        \toprule[1.5pt]
        \textbf{Model} & \textbf{R1} & \textbf{R2} & \textbf{RL}\\
        \midrule[0.5pt]
        \addlinespace[0.35em]
        Other Abs. Sum. models\textsuperscript{*}\\
        \addlinespace[0.5em]
        \hspace{0.5em} \citet{celikyilmaz2018deep} & 41.69 & 19.47 & 37.92\\
        \hspace{0.5em} CopyTransformer (4-layer) & 39.25 & 17.54 & 36.45\\
        \hspace{1.2em} \citet{gehrmann2018bottom} \\
        \hspace{0.5em} GPT-2 (48-layer, zero-shot)& 29.34 & \phantom{0}8.27 & 26.58\\
        \hspace{1.2em} \citet{radford2019language} \\
        \addlinespace[0.75em]        
        No Pre-training\\
        \addlinespace[0.5em]
        \hspace{0.5em} BidirEncoder-Decoder (4-layer) & 37.74 & 16.27 & 34.76\\
        \hspace{0.5em} Encoder-Decoder (12-layer) & 36.72 & 15.22 & 33.84\\
        \hspace{0.5em} Transformer LM (12-layer) & 37.72 & 16.14 & 34.62\\
        \addlinespace[0.75em]
        With Pre-training (all 12-layer)\\
        \addlinespace[0.5em]
        \hspace{0.5em} Pre-train Encoder only & 36.05 & 15.48 & 33.48\\
        \hspace{0.5em} Pre-train Decoder only & 27.48 & \phantom{0}6.87 & 25.40\\
        \hspace{0.5em} Encoder-Decoder & 39.18 & 17.00 & 36.33\\
        \hspace{0.5em} Transformer LM & 39.65 & 17.74 & 36.85\\
        \addlinespace[0.15em]
        \bottomrule[1.5pt]
    \end{tabular}
    \caption{Summarization results when using the full training set. Our scores are averaged over three models trained with different random seeds. \textsuperscript{*}Other abstractive summarization model scores are provided to contextualize performance on this task but are not directly comparable to our models.}
    \label{table:sumresults}
\end{table}

\paragraph{Training Details.} 
We use the neural architecture and hyperparameters from \citet{radford2018improving}, which have demonstrated excellent results on a variety of tasks.
It is a 12-layer Transformer with 135M parameters.
The same architecture is used for both components in our encoder-decoder models.
For LM pre-training, we use a byte pair encoding vocabulary of 63,807 subwords \cite{sennrich2016neural} and train the model on WikiLM for 30 epochs. 
It converges to a perplexity of 20.5, similar to the 18.4 perplexity reached by \citet{radford2018improving}.\footnote{However, the perplexities are not directly comparable because they are on different corpora}
For supervised training, we found it beneficial to use a lower learning rate ($5\times 10^{-5}$ instead of $2\times 10^{-4}$) and train for fewer epochs (6 instead of 12), when incorporating pre-trained weights, compared to when training from scratch.
During inference, we use beam search with beam size 2 while generating summaries.\footnote{All code available at \url{https://github.com/tensorflow/tensor2tensor}}


\paragraph{Results using the full training set.}
Table~\ref{table:sumresults} shows ROUGE scores for our models with and without pre-training. 
We find that pre-training improves performance by about 2 ROUGE points, on average.
Surprisingly, when only the decoder is pre-trained, ROUGE gets substantially worse.
We speculate this is because the model starting out with a well-trained decoder and poor encoder learns to overly rely on its language modeling abilities and not adequately incorporate information from the encoder.
The Transformer LM outperforms corresponding models both with and without pre-training, despite having almost half as many parameters.
Our best model performs competitively with existing models on the CNN/Daily Mail abstractive summarization task, despite the absence of model augmentations such as a copy mechanism and reinforcement learning objectives.

\subsection{Sample Efficiency Results}
\label{sec:sample_efficiency}

We test the sample efficiency of our summarization models 
by training them on randomly generated 1\%, 2\%, 5\%, 10\%, 20\%, and 50\% subsets of the CNN/Daily Mail training data (287,227 examples), using the same subsets for all models.
To strengthen the baseline encoder-decoder network, we make the encoder bidirectional and use a smaller 4-layer model.
We make no changes to the other hyperparameters except increasing the number of training epochs. 
Figure~\ref{fig:sample_efficiency} illustrates sample efficiency for our encoder-decoder models and Transformer LM, both with and without pre-training. We only report ROUGE-2 scores which are lowest of the three metrics used, though trends are consistent for ROUGE-1 and ROUGE-L.

\begin{table}
    \centering
    \small
    \setlength{\tabcolsep}{0.4em}
    \begin{tabular}{p{0.47\textwidth}}
        \toprule[1.5pt]
        \textbf{Ground Truth:} A man in suburban Boston is selling snow online to customers in warmer states. For \$89, he will ship 6 pounds of snow in an insulated Styrofoam box.\\
        \midrule[0.5pt]
        \textbf{Encoder-Decoder + Pre-training:} NEW: A snowfall of is forecast for New England. NEW: The Massachusetts-based company hopes to sell more than 30,000 bottles of snow. The company says it will use snow from as far as Canada.\\
        \midrule[0.5pt]
        \textbf{Transformer LM + Pre-training:} Kyle Waring will ship you 6 pounds of Boston-area snow in an insulated Styrofoam box -- enough for 10 to 15 snowballs, he says. But not if you live in New England or surrounding states.\\
        \bottomrule[1.5pt]
    \end{tabular}
    \caption{Sample outputs after training on 1\% data. See the supplementary materials for more outputs.}
    \label{table:output}
\end{table}

Unsurprisingly pre-training improves sample efficiency, but gains are much larger when also using the Transformer LM.
Fine-tuning the pre-trained Transformer LM on 1\% data, less than 3,000 examples, results in a model that achieves a ROUGE-2 score of 13.1, compared to the pre-trained Transformer encoder-decoder model which only scores 2.3 ROUGE-2 points.
To ensure that augmenting the model with a copy mechanism does not close the sample efficiency gap, we trained the 4-layer Transformer + coverage + copy mechanism model from \citet{gehrmann2018bottom} on 1\% and 5\% of the summarization data and found that while it slightly outperforms our baseline models (getting 2.5 ROUGE-2 at 1\% and 5.1 ROUGE-2 at 5\%), it still performs much worse than our pre-trained Transformer LM.
Overall, our results indicate that while pre-training does improve sample efficiency, having every parameter being pre-trained instead of only a subset of them provides large gains in limited data settings.

%% file: discussion.tex

\begin{figure}[t]
\includegraphics[width=0.47\textwidth]{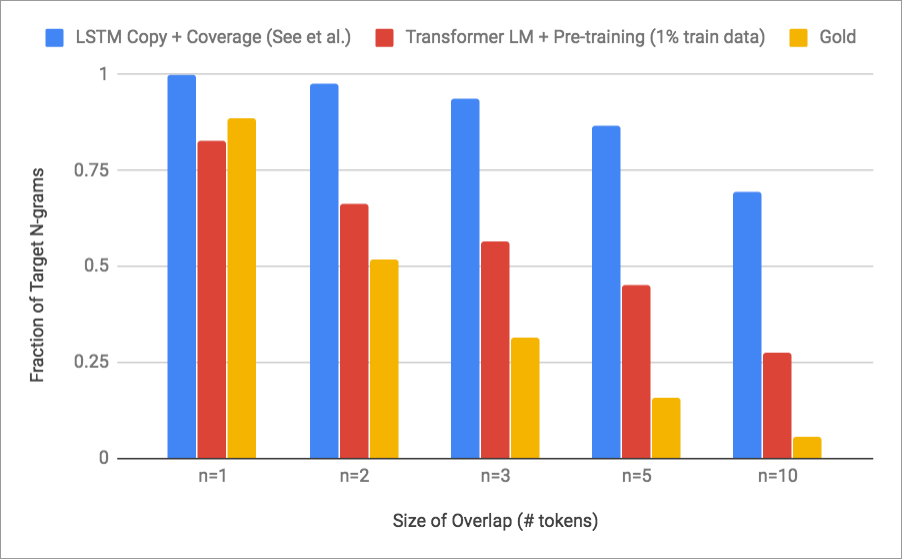}
\caption{$n$-gram overlaps of predicted and gold summaries with their corresponding source articles.
While our model copies more than gold summaries, it copies substantially less than the pointer-generator model.
}
\label{fig:overlaps}
\end{figure}

\section{Analysis}
\label{sec:disc}

Table~\ref{table:output} shows generated outputs for models fine-tuned on 1\% data.
The pre-trained Transformer LM succeeds in copying salient information from the source, which indicates it can effectively attend over the source article.
On the other hand, the pre-trained Encoder-Decoder model hallucinates facts such as ``30,000 bottles of snow", which are topical but never appear in the source, suggesting that the model is unable to utilize information from the source.
Instead, it behaves more as a general domain language model.

The pre-trained Transformer LM's predictions initially raised the concern: how much of the impressive sample efficiency is due to copying from the source? 
After all, the no-training-required ``lead-3" baseline \citep{see2017get}, which uses the first three sentences of the article as the summary, achieves 17.7 ROUGE-2.
Hence, we investigate the extent to which summaries are copied from the articles by computing $n$-gram overlaps between the articles and summaries.
In Figure~\ref{fig:overlaps}, we compare $n$-gram overlaps for (1) our pre-trained Transformer LM fine-tuned on 1\% data, (2) \citet{see2017get}'s pointer-generator with coverage model, and (3) gold summaries. While our model copies from the source more often than gold summaries, it is more abstractive than the pointer-generator model which copies 70\% of all generated 10-grams, compared to our model which copies 27\% of its generated 10-grams, likely due to lack of a copy mechanism.

%% file: conclusion.tex
\section{Conclusion}
\label{sec:conc}

Sample efficiency can be vital for narrow domains and low-resource settings, especially in the case of generation tasks for which models often require large datasets to perform well.
In this paper, we have shown that using a single pre-trained Transformer LM for sequence-to-sequence tasks simplifies the model, reduces the number of parameters, and removes the non-pre-trained encoder-decoder attention weights.
More importantly, experiments fine-tuning the model on only 1\% training data have shown that our approach achieves impressive sample efficiency gains. 
It would be interesting to further test whether this approach leads to similar sample efficiency gains on tasks beyond summarization, such as dialogue.

%% file: appendix.tex
\newpage
\section{Appendices}
\label{sec:appendix}

\definecolor{c0}{rgb}{0.8,0,0}
\definecolor{c1}{rgb}{0,0.0,0.8}
\definecolor{c2}{rgb}{0.0,0.7,0.0}
\definecolor{c3}{rgb}{1.0,0.0,0.5}
\definecolor{c4}{rgb}{0.0,0.5,1.0}

\setlength{\tabcolsep}{1mm}
\begin{table*}
\small
\begin{tabularx}{\textwidth}{X X X}
\toprule[1.5pt]
\makecell{Encoder-Decoder + Pre-Training  \\ Fine-tuned on \textbf{1\%} of training data} &
\makecell{Transformer LM + Pre-Training \\ Fine-tuned on \textbf{1\%} of training data} &
\makecell{Pointer-Generator \citep{see2017get} \\ Trained on \textbf{100\%} of training data}\\
\midrule[0.5pt]

\color{c0}[the] \color{black}[u.s.] \color{black}[u.s.] \color{black}[u.s] \color{c1}[. the] \color{black}[u.s] \color{c2}[.] \color{c3}[it] \color{black}[has] \color{black}[been] \color{c4}[in the] \color{black}[u.s] \color{c0}[. the] \color{black}[u.s] \color{c1}[.] \color{c2}[it] \color{c3}[is] \color{c4}[the] \color{black}[u.s] \color{c0}[. the] \color{c1}[u] \color{c2}[.]   &  \color{c0}[president] \color{c1}[obama says he is `` absolutely committed] \color{c2}[''] \color{c3}[to making sure] \color{c4}[israel is] \color{c0}[not] \color{c1}[vulnerable] \color{c2}[.] \color{c3}[netanyahu] \color{c4}[says he] \color{c0}[understands and respects] \color{c1}[obama] \color{c2}['s stance that israel is particularly vulnerable] \color{c3}[.]   &  \color{c0}[obama says he is `` absolutely committed to making sure '' israel maintains a military advantage over iran . his comments to the new york times] \color{c1}[published on sunday , come amid criticism from israeli prime minister benjamin netanyahu] \color{c2}[.] \\
\color{c0}[new] \color{black}[:] \color{c1}[the] \color{black}[president] \color{black}[will] \color{c2}[be] \color{c3}[able] \color{c4}[to the] \color{black}[world] \color{c0}['s] \color{c1}[new] \color{c2}[new] \color{c3}[new] \color{black}[president] \color{c4}[. the group] \color{c0}[says] \color{c1}[it] \color{black}[will] \color{c2}[be] \color{c3}[able] \color{c4}[to] \color{c0}[be] \color{c1}[able .] \color{c2}[the] \color{black}[u.s] \color{c3}[. the] \color{black}[u.s] \color{c4}[. the] \color{black}[u.s] \color{c0}[.] \color{c1}[it] \color{black}[will] \color{c2}[be] \color{c3}[used] \color{c4}[to] \color{c0}[be] \color{c1}[able] \color{c2}[to] \color{c3}[be] \color{c4}[able] \color{c0}[to] \color{c1}[be] \color{c2}[able .] \color{c3}[the u] \color{c4}[.]   &  \color{c0}[the american pharmacists association] \color{c1}[voted at its annual meeting to adopt a ban as an official policy] \color{c2}[.]   &  \color{c0}[the american pharmacists association is discouraging its members from participating in executions .] \color{c1}[the group acted this week because of increased public attention on lethal injection] \color{c2}[. thirty-two sates allow capital punishment , and lethal injection is still the most common method .] \\
\color{c0}[new] \color{black}[:] \color{c1}[the] \color{black}[president] \color{c2}[is] \color{c3}[the] \color{black}[world] \color{c4}['s] \color{black}[death] \color{c0}[of the] \color{black}[world] \color{c1}['s] \color{black}[first] \color{c2}[time] \color{c3}[at the] \color{black}[world] \color{c4}['s] \color{black}[death] \color{c0}[. the] \color{black}[president] \color{c1}[was] \color{black}[found] \color{c2}[in the] \color{black}[world] \color{c3}['s] \color{black}[first] \color{c4}[time] \color{c0}[of the] \color{black}[world] \color{c1}['s] \color{black}[first] \color{c2}[time] \color{c3}[of the] \color{black}[world] \color{c4}['s] \color{black}[first] \color{c0}[time] \color{c1}[of the] \color{black}[world] \color{c2}[.]   &  \color{c0}[zaki-ur-rehman lakhvi , a top leader of] \color{c1}[lashkar-e-taiba , was released early friday from] \color{c2}[jail in the pakistani city of rawalpindi] \color{c3}[.] \color{c4}[he] \color{c0}[is] \color{c1}[accused of masterminding the november 2008 terror attacks that left more than 160 people dead in mumbai , india 's most populous city .]   &  \color{c0}[zaki-ur-rehman lakhvi , a top leader of the terrorist group lashkar-e-taiba , was released early friday from a jail in the pakistani city of rawalpindi] \color{c1}[. lakhvi was charged in pakistan in 2009 , accused of masterminding the november 2008 terror attacks that left more than 160 people dead in mumbai , india 's most populous city .] \\
\color{c0}[new] \color{c1}[:] \color{c2}[the] \color{c3}[president] \color{c4}[is] \color{c0}[the] \color{c1}[world] \color{c2}['s] \color{c3}[president] \color{c4}[president] \color{c0}[is] \color{c1}[the] \color{c2}[world] \color{c3}['s] \color{c4}[first] \color{c0}[time] \color{c1}[of the] \color{c2}[world] \color{c3}['s] \color{c4}[most] \color{c0}[most] \color{c1}[most] \color{c2}[most] \color{c3}[of the] \color{c4}[world] \color{c0}[.] \color{c1}[he] \color{black}[says] \color{c2}[he] \color{c3}['s] \color{c4}[first] \color{c0}[time] \color{c1}[to the] \color{c2}[world] \color{black}[cup] \color{black}[final] \color{c3}[.]   &  \color{c0}[don mclean 's] \color{c1}[`` american pie '' is] \color{c2}[44 years old .] \color{c3}[he] \color{c4}[was a paperboy when , on february 3 , 1959 , he saw that buddy holly , ritchie valens and j.p. `` the big bopper '' richardson had been tragically killed in an airplane crash in clear lake , iowa .]   &  \color{c0}[don mclean 's pop masterpiece `` american pie] \color{c1}['' is] \color{c2}[44 years old .] \color{c3}[christie 's sold the 16-page handwritten manuscript of the song 's lyrics for \$ 1.2 million to an unnamed buyer . mclean was a paperboy when , on february 3 , 1959 , he saw that buddy] \color{c4}[valens and j.p. `` the big bopper '' richardson had been tragically killed] \color{c0}[.] \\
\color{c0}[new] \color{c1}[: the] \color{c2}[president] \color{c3}[is] \color{c4}[the world] \color{c0}['s] \color{black}[first] \color{c1}[time] \color{c2}[of the] \color{c3}[world] \color{c4}['s] \color{black}[first] \color{c0}[time] \color{c1}[at] \color{c2}[the world] \color{c3}['s] \color{black}[most] \color{black}[most] \color{black}[most] \color{black}[most] \color{c4}[. the] \color{c0}[president] \color{c1}[is] \color{c2}[the] \color{black}[first] \color{c3}[time] \color{c4}[of the] \color{c0}[world] \color{c1}['s] \color{black}[most] \color{black}[most] \color{black}[most] \color{black}[most] \color{black}[most] \color{black}[most] \color{black}[most] \color{c2}[of the] \color{c3}[world] \color{c4}[.]   &  \color{c0}[iran] \color{c1}[won the world cup] \color{c2}[after a] \color{c3}[nuclear deal with the united states] \color{c4}[. the] \color{c0}[deal] \color{c1}[promises to end iran 's international isolation under years of crippling sanctions .]   &  \color{c0}[iranians erupted in celebration as young people waved flags from their sunroofs , blasted music from stereos and chatted online with the hashtag \# irantalks .] \color{c1}[excitement came after a breakthrough nuclear deal with the united states and other world powers that promises to end iran 's international isolation under years of crippling sanctions .] \\
\color{c0}[new] \color{black}[:] \color{c1}[a] \color{black}[woman] \color{c2}[has] \color{black}[been] \color{black}[charged] \color{c3}[with] \color{c4}[a] \color{c0}[year] \color{c1}[.] \color{c2}[she] \color{black}[says] \color{c3}[it 's] \color{black}[first] \color{black}[time] \color{c4}[to be able to] \color{c0}[be able] \color{c1}[. the] \color{black}[woman] \color{c2}['s new] \color{black}[woman] \color{c3}[is] \color{c4}[a] \color{black}[woman] \color{c0}['s] \color{black}[most] \color{c1}[of the] \color{black}[woman] \color{c2}['s new] \color{c3}[new] \color{black}[woman] \color{c4}[is] \color{c0}[in the] \color{black}[first] \color{black}[time] \color{c1}[.]   &  \color{c0}[netflix] \color{c1}[ordered up a reunion special , followed by a spinoff series called `` fuller house] \color{c2}['' the show will] \color{c3}[be available next year , netflix said .]   &  \color{c0}[john stamos announced monday night on `` jimmy kimmel live ''] \color{c1}[. the show will] \color{c2}[feature candace cameron bure , who played eldest daughter d.j . tanner in the original series] \color{c3}[,] \color{c4}[which aired from 1987 to 1995] \color{c0}[, will both return for the new series] \color{c1}[.]

 \\
\bottomrule[1.5pt]
\end{tabularx}
\caption{Comparisons of summaries generated by various models. Colors/brackets correspond to consecutive words that occur in the article (black means the word was not in the article text).}

\label{tab:summary_comparisons}
\end{table*}

Our Transformer LM + pre-training system is extremely data-efficient: it generates decent summaries using only 1\% of the training data while the baseline pre-trained Encoder-Decoder model essentially generates gibberish.
Examples illustrating this behavior are shown in Table~\ref{tab:summary_comparisons}.
Furthermore, our Transformer LM appears to be more abstractive than the Pointer-Generator network, which uses a copy mechanism.
While the Pointer-Generator often copies over whole sentences from the article, our model mixes and matches sentence fragments, with other generated words sometimes connecting them. The downside of this abstractiveness is the way it occasionally ``hallucinates" facts, such as in the amusing summary in row 5.